\documentclass{article}

\usepackage{amsmath}
\usepackage{microtype}
\usepackage[T1]{fontenc}
\usepackage{graphicx}
\usepackage{subcaption}
\usepackage{booktabs} 

\usepackage{hyperref}

\newcommand{\btvpp}{\texttt{baller2vec++}}

\usepackage[accepted]{innf2021}

\icmltitlerunning{The DEformer: An Order-Agnostic Distribution Estimating Transformer}

\begin{document}

\twocolumn[
\icmltitle{The DEformer: An Order-Agnostic Distribution Estimating Transformer}



\icmlsetsymbol{equal}{*}

\begin{icmlauthorlist}
\icmlauthor{Michael A. Alcorn}{au}
\icmlauthor{Anh Nguyen}{au}
\end{icmlauthorlist}

\icmlaffiliation{au}{Department of Computer Science and Software Engineering, Auburn University, Auburn, Alabama, USA}

\icmlcorrespondingauthor{Michael A. Alcorn}{alcorma@auburn.edu}

\icmlkeywords{Machine Learning, ICML}

\vskip 0.3in
]



\printAffiliationsAndNotice{}  

\begin{abstract}
\setcounter{footnote}{1}
Order-agnostic autoregressive distribution (density) estimation (OADE), i.e., autoregressive distribution estimation where the features can occur in an arbitrary order, is a challenging problem in generative machine learning.
Prior work on OADE has encoded feature identity by assigning each feature to a distinct fixed position in an input vector.
As a result, architectures built for these inputs must strategically mask either the input or model weights to learn the various conditional distributions necessary for inferring the full joint distribution of the dataset in an order-agnostic way.
In this paper, we propose an alternative approach for encoding feature identities, where each feature's identity is included \textit{alongside} its value in the input.
This feature identity encoding strategy allows neural architectures designed for sequential data to be applied to the OADE task without modification.
As a proof of concept, we show that a Transformer trained on this input (which we refer to as ``the DEformer''\footnote{All data and code for the paper are available at:  \url{https://github.com/airalcorn2/deformer}.}, i.e., the distribution estimating Transformer) can effectively model binarized-MNIST, approaching the performance of fixed-order autoregressive distribution estimating algorithms while still being entirely order-agnostic.
Additionally, we find that the DEformer surpasses the performance of recent flow-based architectures when modeling a tabular dataset.
\end{abstract}

\section{Introduction}
For tasks such as: (a) efficiently imputing arbitrary missing values from an input or (b) preemptive anomaly detection in systems where input features can arrive asynchronously in an arbitrary order (e.g., internet of things applications \cite{ahmad2017unsupervised}), order-agnostic autoregressive distribution (density) estimation (OADE) is necessary.
However, because there are $D!$ factorizations of the joint probability for a $D$-dimensional input, order-agnosticism adds considerable complexity to the distribution estimation task.
As a result, many likelihood-based generative models either: (1) assume a single, fixed order for the input features (e.g., NADE \cite{larochelle2011neural}, PixelRNN \cite{oord2016pixelrnn}, and TraDE \cite{fakoor2020trade}), (2) only use a small subset of the possible feature orderings in practice (e.g., MADE \cite{germain2015made}, IAF \cite{kingma2016iaf}, MAF \cite{papamakarios2017maf}, and LMConv \cite{jain2020lmconv}), or (3) are not autoregressive (e.g., some flows \cite{DBLP:journals/corr/DinhKB14, dinh2017realNVP, kingma2018glow, Papamakarios2021normalizing}).

In contrast to the previously mentioned approaches, DeepNADE \cite{uria2014deep, uria2016neural} is notable in that it performs full OADE.
Specifically, DeepNADE consists of a standard multilayer perceptron (MLP) that takes as input the concatenation of a $D$-dimensional binary mask $\textbf{m}$ and the masked version of the sample $\textbf{x}$, $\hat{\textbf{x}} = \textbf{m} \odot \textbf{x}$, i.e., the input $[\hat{\textbf{x}}, \textbf{m}]$ is a vector of size $2D$. 
The feature identities (e.g., pixel locations) are thus encoded by their positions in the input feature vectors.
However, this input design precludes the use of neural architectures that are designed for sequential data (e.g., recurrent neural networks and Transformers \cite{vaswani2017attention})—models that are a natural fit for autoregressive problems.

Taking inspiration from a recently described multi-agent spatiotemporal Transformer \cite{alcorn2021baller2vecplusplus}, in this paper, we propose an alternative approach for encoding feature identities, where each feature's identity is included \textit{alongside} its value in the input.
Using this input design, we train an otherwise ordinary Transformer (which we refer to as ``the DEformer'', i.e., the distribution estimating Transformer) to perform OADE on the binarized-MNIST \cite{salakhutdinov2008quantitative} and POWER \cite{vergara2012chemical} datasets.
We find that:

\begin{enumerate}
    \item The DEformer—while being entirely order agnostic and autoregressive—is competitive with fixed-order distribution estimating algorithms when modeling binarized-MNIST and surpasses recent flow-based architectures when modeling the tabular POWER dataset.
    \item The DEformer can effortlessly fill in pixels of binarized-MNIST images that are missing in a variety of patterns.
    \item The DEformer can easily distinguish between binarized-MNIST and non-binarized-MNIST images.
\end{enumerate}

\section{Architecture}
\begin{figure}[h]
\centering
\includegraphics[width=\columnwidth]{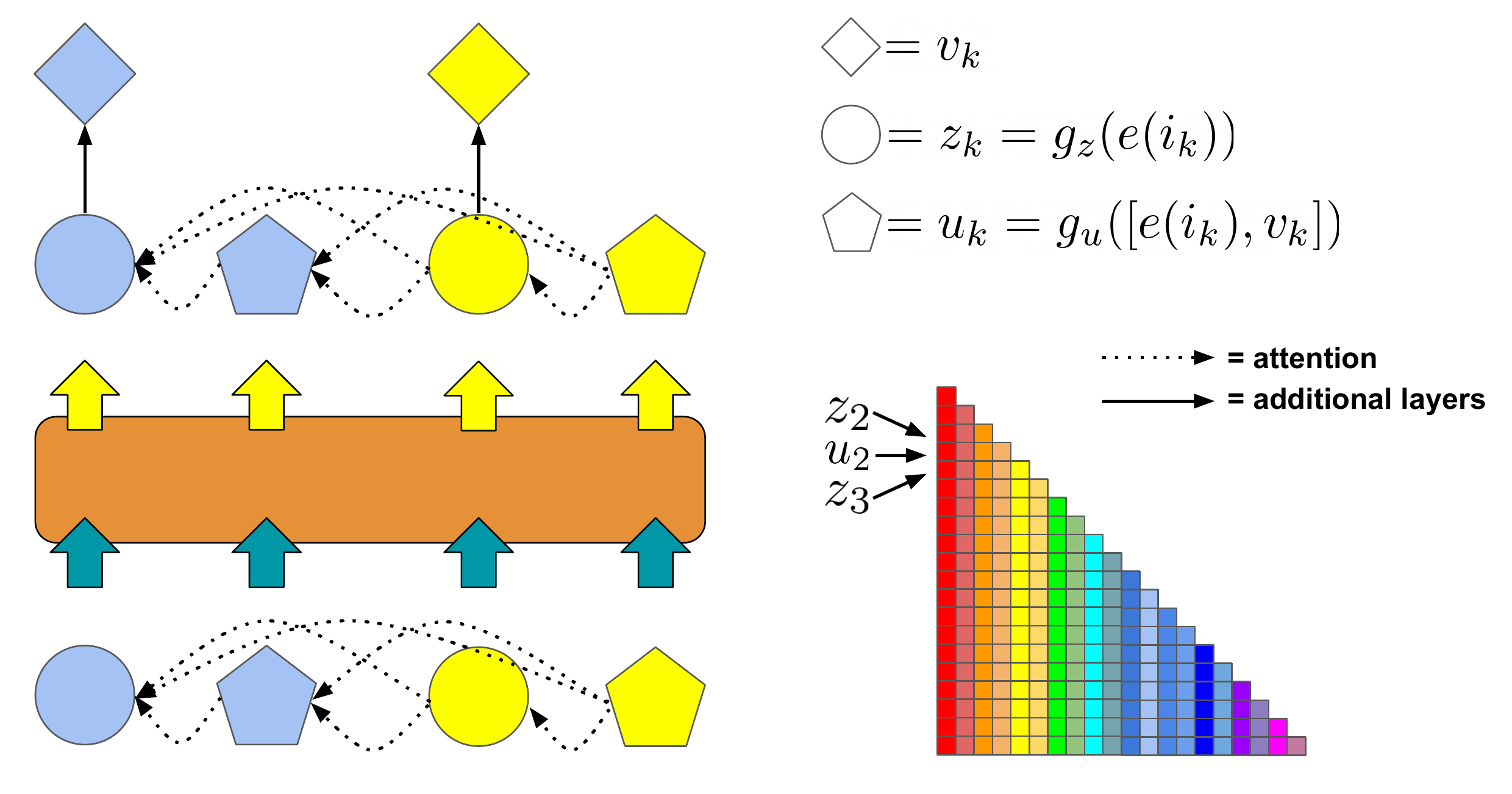}
\caption{By including each feature's identity \textit{alongside} its value in the input, sequential models can be used to perform order-agnostic autoregressive distribution estimation.
The DEformer is a Transformer that uses an interleaved input design (partially depicted here with the self-attention mask) for this task.
The two sets of interleaved feature vectors consist of identity feature vectors ($z_{k}$) and identity/value feature vectors ($u_{k}$), and $g_{z}$ and $g_{u}$ are their respective multilayer perceptrons.
For the binarized-MNIST dataset, each feature identity $i_{k}$ is a tuple ($r_{k}$, $c_{k}$) where $r_{k}$ and $c_{k}$ are the row and column for the pixel indexed by $k$ in the permuted sequence, respectively, and $v_{k}$ is the value of the pixel (which is zero or one for binary images).
For tabular data, each $i_{k}$ corresponds to a column, and $v_{k}$ is the value of the indexed column in the row.
Lastly, $e$ is an identity encoding function, which is simply the identify function in the case of binarized-MNIST, and is an embedding layer for tabular data.
}
\label{fig:deformer}
\end{figure}

Here, we describe our order-agnostic distribution estimating Transformer, the DEformer (Figure \ref{fig:deformer}).
The goal in OADE is to model the joint distribution of a $D$-dimensional vector $\textbf{x}$ by exploiting the chain rule of probability, i.e.:

\[
    p(\textbf{x}) = \Pi_{d = 1}^{D}p(x_{o_{d}} | \textbf{x}_{o_{< d}})
\]

\noindent
where, as in \citet{uria2014deep}, $o$ is a $D$-tuple representing a permutation of the elements in $\textbf{x}$, so $x_{o_{d}}$ indicates the element of $\textbf{x}$ indexed by the $d$-th element of $o$, and $\textbf{x}_{o_{< d}}$ means the elements in $\textbf{x}$ indexed by the first $d - 1$ elements of $o$.
We assume each discrete feature can take on one of $C$ labels (which is the case for image datasets), but, in theory, each feature could have a different number of possible labels.

Rather than encoding each feature's identity by confining it to a specific position in the input, here, we propose including the feature's identity as an additional input variable \textit{alongside} its value.
Specifically, the input to the DEformer consists of two parallel sequences: one containing only feature identities, and another containing identity/value pairs:

\begin{enumerate}
    \item $i_{1}, i_{2}, ..., i_{n}$
    \item $(i_{1}, v_{1}), (i_{2}, v_{2}), ..., (i_{n}, v_{n})$
\end{enumerate}

\noindent
where $i_{k}$ is the identity of the $k$-th feature in the permuted sequence and $v_{k}$ is the value of the $k$-th feature.
In the case of binarized-MNIST, each $i_{k}$ is a tuple $(r_{k}, c_{k})$ indicating the row and column of the pixel, respectively, and $v_{k}$ is the value of the pixel (i.e., zero or one).
For tabular data, each $i_{k}$ indexes a column, and $v_{k}$ is the value of the indexed column in the row.

The identity inputs are mapped to identity feature vectors using an MLP, i.e., $z_{k} = g_{z}(e(i_{k}))$ where $z_{k}$ is the identity feature vector, $g_{z}$ is the identity MLP, and $e$ is an identity encoding function.
In the case of binarized-MNIST, $e$ is simply the identity function, i.e., $e(i_{k}) = [r_{k}, c_{k}]$, while for tabular data, $e$ is an embedding layer.
The identity/value pairs are similarly mapped to identity/value feature vectors using a separate MLP, i.e., $u_{k} = g_{u}([e(i_{k}), v_{k}])$ where $u_{k}$ is the identity/value feature vector and $g_{u}$ is the identity/value MLP.
These two sets of feature vectors are interleaved with one another (i.e., $u_{k}$ always immediately follows $z_{k}$ in the input) to form a $2D \times F$ matrix where $F$ is the dimension of the outputs for the MLPs.

This matrix is passed into the Transformer along with a lower triangular self-attention mask, which encodes the following dependencies (see Figure \ref{fig:deformer}):

\begin{enumerate}
    \item When processing $z_{k_{2}}$, the DEformer is allowed to ``look'' at: (i) any $z_{k_{1}}$ where $k_{1} \leq k_{2}$ and (ii) any $u_{k_{1}}$ where $k_{1} < k_{2}$.
    \item When processing $u_{k_{2}}$, the DEformer is allowed to ``look'' at: (i) any $z_{k_{1}}$ where $k_{1} \leq k_{2}$ and (ii) any $u_{k_{1}}$ where $k_{1} \leq k_{2}$.
\end{enumerate}

\noindent
Like \citet{alcorn2021baller2vec, alcorn2021baller2vecplusplus}, we do not use positional encoding \cite{vaswani2017attention} because \citet{Irie2019} observed that positional encoding is not only unnecessary, but detrimental for Transformers that use a causal attention mask.

Each processed $z_{k}$ feature vector is then passed through a final linear layer.
When modeling discrete features, the final linear layer is followed by a softmax, which gives a probability distribution over the labels for the feature indexed by $k$. The loss for each sample is thus:

\begin{equation}\label{eq:discrete_loss}
    \mathcal{L} = \sum_{k=1}^{K} -\ln(f(Z)_{2k - 1}[v_{k}])
\end{equation}

\noindent
where $f(Z)_{2k - 1}[v_{k}]$ is the probability assigned to the label $v_{k}$ (where $v_{k}$ is an integer from one to $C$) by $f$, i.e., Equation \eqref{eq:discrete_loss} is the NLL of the data according to the model.
For continuous features, the output of the final linear layer defines a mixture of Gaussians, so the loss for each sample is:

\[
    \mathcal{L} = \sum_{k=1}^{K} -\ln(\pi_{k} \cdot c_{k})
\]

\noindent
where $\pi_{k} = \text{softmax}(f(Z)_{2k - 1,1:J})$ is a vector containing the $J$ mixture proportions for feature $k$, and $c_{k}$ is a vector containing the mixture densities such that:

\[
    c_{k}[j] = \frac{1}{\sigma_{k,j} \sqrt{2 \pi}} e^{-\frac{1}{2} (\frac{v_{k} - \mu_{k,j}}{\sigma_{k,j}})^{2}}
\]

\noindent
where $\sigma_{k} = f(Z)_{2k - 1,J + 1:2J}$ and $\mu_{k} = f(Z)_{2k - 1,2J + 1:3J}$.

Because any ordering of a chain rule decomposition of a joint probability produces the same value, e.g.:
\vskip -0.2in

\[
    p(x_{1}) p(x_{2} | x_{1}) p(x_{3} | x_{1} x_{2}) = p(x_{3}) p(x_{2} | x_{3}) p(x_{1} | x_{3} x_{2})
\]

\noindent
like \citet{uria2014deep, yang2019xlnet, alcorn2021baller2vecplusplus}, we shuffle the order of the features in each training sample to encourage the DEformer to learn a joint distribution of the dataset that is approximately permutation invariant with respect to the ordering of the features.

\section{Experiments}
To test the utility of the DEformer for OADE, we trained a nearly identical architecture to the model described in \citet{alcorn2021baller2vecplusplus} on the binarized-MNIST \cite{salakhutdinov2008quantitative} and POWER \cite{vergara2012chemical} datasets.
The binarized-MNIST dataset consists of 70,000 $28 \times 28$ pixel binary images (i.e., the pixel values are either black or white) of digits (i.e., 0-9) where each digit is represented by the same number of images.
We used the standard 60,000/10,000 split for training/testing images, respectively, and used 1,200 of the 60,000 training images (i.e., 2\%) for validation.
The POWER dataset consists of 2,049,280 power measurements from a single household in a tabular format, where each sample consists of six real values.
We used the same preprocessing steps and training/validation/test split described in \citet{papamakarios2017maf}.

The size of the output for the final linear layer was one for the binarized-MNIST dataset and $3 \times 150 = 450$ for the POWER dataset (as in \citet{fakoor2020trade}), but all remaining hyperparameters and training details were nearly identical to \btvpp{} \cite{alcorn2021baller2vecplusplus}, which itself closely follows the original Transformer \cite{vaswani2017attention}.
Specifically, the Transformer settings were: $d_{\text{model}} = 512$ (the dimension of the input and output of each Transformer layer), eight attention heads, $d_{\textrm{ff}} = 2048$ (the dimension of the inner feedforward layers), six layers, dropout probabilities of 0.0 and 0.2 for the binarized-MNIST and POWER datasets, respectively, and no positional encoding.
Each MLP (i.e., $g_{z}$, $g_{u}$, and $g_{r}$) had 128, 256, and 512 nodes in its three layers, respectively, and a ReLU nonlinearity following each of the first two layers.
Lastly, the identity embedding layer for the POWER dataset mapped column indices to 20-dimensional vectors.

\begin{figure}[t]
\centering
\begin{subfigure}{\columnwidth}
    \centering
    \includegraphics[width=0.75\columnwidth]{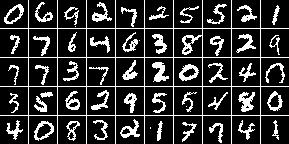}
\end{subfigure}
\vskip 0.2in
\begin{subfigure}{\columnwidth}
    \centering
    \includegraphics[width=0.75\columnwidth]{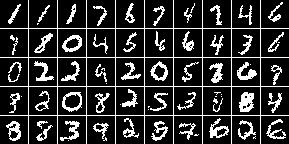}
\end{subfigure}
\caption{\textbf{Top}: A sample of 50 images from the test set of binarized-MNIST organized by their average NLLs according to the DEformer (starting with the lowest average NLL, 42.1, at the top left, and ending with the highest average NLL, 119.6, at the bottom right).
\textbf{Bottom}: 50 generated images organized by their average NLLs according to the DEformer (starting with the lowest average NLL, 42.0, at the top left, and ending with the highest average NLL, 131.1, at the bottom right).
The average NLLs for both sets of images are calculated over 10 random orderings.
}
\label{fig:test_nlls}
\vskip -0.1in
\end{figure}

We used the Adam optimizer \cite{kingma2014adam} with an initial learning rate of $10^{-6}$, $\beta_{1} = 0.9$, $\beta_{2} = 0.999$, and $\epsilon = 10^{-9}$ to update the model parameters, of which there were $\sim$19 million.
The learning rate was reduced to $10^{-7}$ after 5/20 epochs of the validation loss not improving for the binarized-MNIST/POWER datasets, respectively, and we used batch sizes of 1/128 for the binarized-MNIST/POWER datasets, respectively.
Models were implemented in PyTorch and trained on a single NVIDIA GTX 1080 Ti GPU for $\sim$50/700 epochs (2.5/6 days) for the binarized-MNIST/POWER datasets, respectively, and the validation set was used for early stopping.

\begin{table}[t]
\caption{The average NLL on the binarized-MNIST test set for different models.
Despite being entirely order agnostic (``OA''), the DEformer is competitive with PixelRNN and TraDE, which use a single fixed order (``FO'').
For MADE, the model was trained on 32 different orders \cite{uria2016neural}.
The average NLLs for both DeepNADE and the DEformer are calculated over 10 random orderings.}
\label{sample-table}
\vskip 0.15in
\begin{center}
\begin{small}
\begin{sc}
\begin{tabular}{lc}
\toprule
Model & NLL \\
\midrule
DeepNADE (OA) & 89.17 \\
MADE (32) & 86.64 \\
PixelRNN (FO) & 79.20 \\
TraDE (FO) & 78.92 \\
DEformer (OA) & 80.49 \\
\bottomrule
\end{tabular}
\end{sc}
\end{small}
\end{center}
\vskip -0.1in
\label{tab:nlls}
\end{table}

\begin{table}[t]
\caption{The average NLL on the POWER test set for different models.
The DEformer surpasses the performance of recent flow-based architectures while still retaining order-agnostic and autoregressive properties.
The average NLL for the DEformer is calculated over 10 random orderings.}
\label{sample-table}
\vskip 0.15in
\begin{center}
\begin{small}
\begin{sc}
\begin{tabular}{lc}
\toprule
Model & NLL \\
\midrule
RealNVP & -0.17 \\
MAF & -0.3 \\
NAF & -0.62 \\
NSF & -0.66 \\
TraDE & -0.73 \\
DEformer & -0.68 \\
\bottomrule
\end{tabular}
\end{sc}
\end{small}
\end{center}
\vskip -0.1in
\label{tab:table_nlls}
\end{table}

\begin{figure}[t]
\centering
\includegraphics[width=0.65\columnwidth]{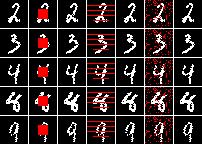}
\caption{Because the DEformer is order-agnostic, it can easily ``fill in'' images where pixels are missing in a variety of patterns by placing the missing pixels at the end of the input sequence.
Here, each row corresponds to a different ground truth image from the test set (depicted in the first column).
The remaining pairs of columns show 100 removed pixels (red) from the ground truth image and the corresponding filled in image.
}
\label{fig:filled}
\end{figure}

\section{Results}
\begin{figure}[h]
\centering
\includegraphics[width=\columnwidth]{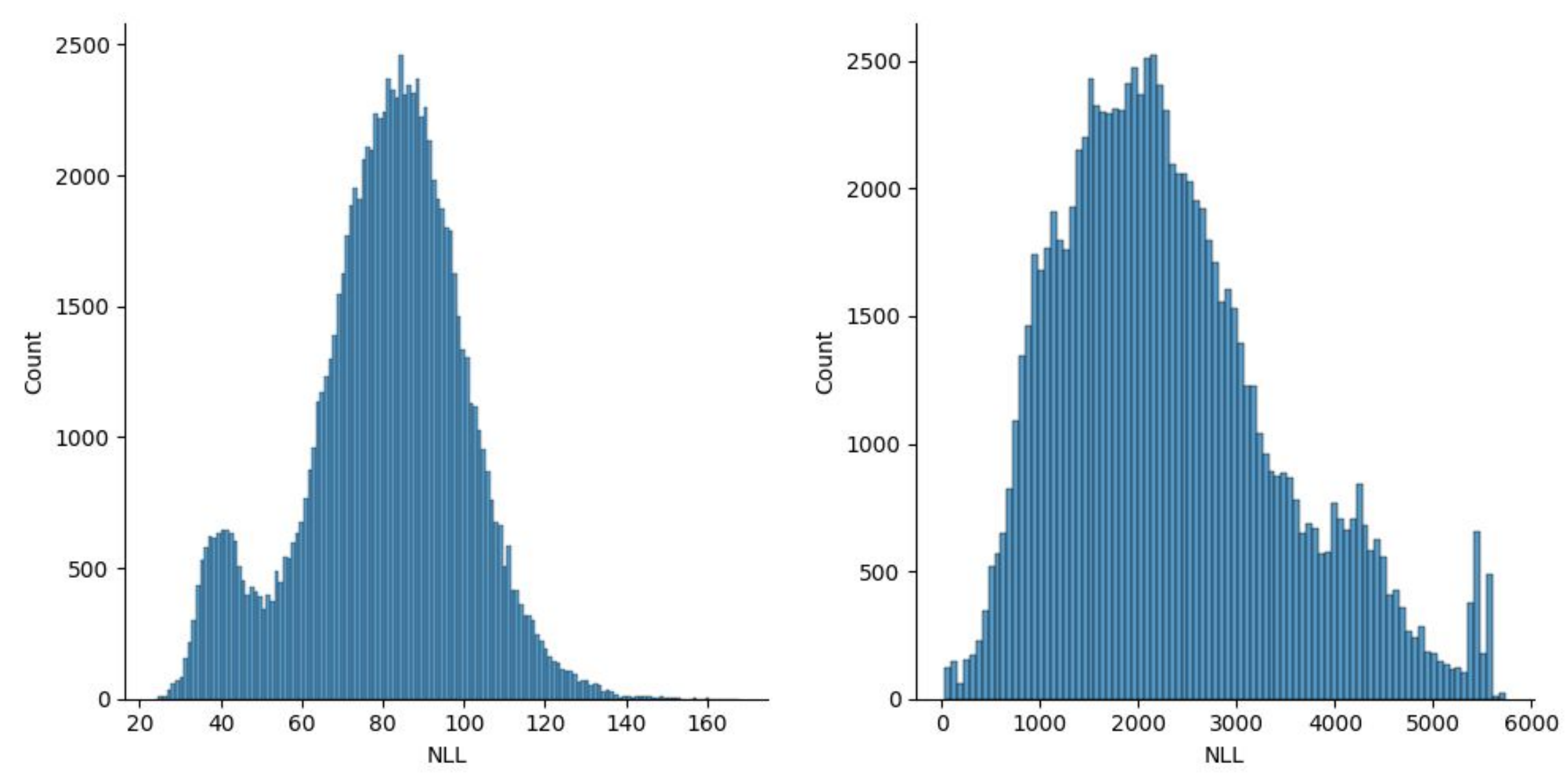}
\caption{The distribution of the DEformer average NLLs for the binarized-MNIST test set (left) and a subset of 10,000 images from the binarized-notMNIST dataset (right) diverge considerably (notice the difference in scales of the $x$-axes), i.e., the DEformer consistently assigns lower probabilities to out-of-distribution samples.
}
\label{fig:notmnist}
\vskip -0.2in
\end{figure}

The DEformer achieved an average NLL (taken over 10 orders) of 80.49 on the binarized-MNIST test set.
This is a vast improvement over DeepNADE \cite{uria2014deep} and is competitive with fixed-order distribution estimation algorithms like PixelRNN \cite{oord2016pixelrnn} and TraDE \cite{fakoor2020trade} (see Table \ref{tab:nlls}).
On the POWER dataset, the DEformer achieved an average NLL of -0.68, which surpasses the performance of recent flow-based architectures like NAF \cite{huang2018neural} and NSF \cite{durkan2019neural}.
We suspect the DEformer's performance could be improved with a careful hyperparameter search.

Following \citet{uria2014deep}, Figure \ref{fig:test_nlls} shows 50 samples from the test set of binarized-MNIST sorted by their average NLLs (taken over 10 orders) according to the DEformer, along with 50 samples generated by the DEformer, also sorted by their average NLLs.
Also following \citet{uria2014deep}, Figure \ref{fig:filled} shows examples of images with 100 pixels missing in a variety of patterns, which were then ``filled in'' by the DEformer when conditioned on the remaining 684 pixels.
Like DeepNADE, this task is trivial for the DEformer because the pixels can be arranged such that the conditioning pixels are at the beginning of the sequence.
Lastly, as can be seen in Figure \ref{fig:notmnist}, the DEformer can easily distinguish between in-distribution and out-of-distribution (i.e., binarized-notMNIST images \cite{bulatov_2011}) samples.

\section{Related Work}
\subsection{Interleaved input Transformers}

The DEformer is directly inspired by \btvpp{} \cite{alcorn2021baller2vecplusplus}, a multi-agent spatiotemporal Transformer that used an identical interleaved input design to model the behaviors of coordinated agents.
Our key contribution is recognizing that this interleaved architecture design can be applied to OADE.
The DEformer is architecturally similar to the independently developed XLNet language model \cite{yang2019xlnet}.
Compared to XLNet, the DEformer:

\begin{enumerate}
    \item encodes feature identity by including it as an input to the network (instead of using positional embeddings) and
    \item uses a full lower triangular attention mask to attend to \textit{both} identity feature vectors and identity/value feature vectors that occur earlier in the shuffled input (instead of only attending to the ``content stream'').
\end{enumerate}

\noindent
Notably, XLNet was trained to only predict the final six tokens of a shuffled sentence because the authors observed ``slow convergence in preliminary experiments''.
The DEformer was capable of modeling the values for all 784 pixels in our binarized-MNIST experiments.

\subsection{DeepNADE}

One important way DeepNADE \cite{uria2014deep, uria2016neural} and the DEformer differ is in the size of the outputs for their final classification layers, which are $DC$ and $C$, respectively.
While this difference is not particularly important for a relatively simple dataset like binarized-MNIST, for more complex datasets like CIFAR-10 \cite{krizhevsky2009learning}, these contrasting designs produce dramatically different parameter counts.
Specifically, the size of the output for a CIFAR-10 DeepNADE model would be $32 \times 32 \times 3 \times 256 =$ 786,432 (because each pixel has three channels, and each channel can take on one of 256 different integer values).
Therefore, if the input dimension to the final layer was 500 (as it was in the DeepNADE model for binarized-MNIST), the final layer alone would have 500 $ \times $ 786,432 + 786,432 $ = $ 394,002,432 parameters.
While the number of outputs can be reduced for image datasets by using a discretized logistic mixture likelihood \cite{Salimans2017PixeCNN}, this strategy restricts the complexity of the model, and the discretized logistic mixture likelihood is not applicable to datasets where the labels do not have a clear underlying order.

On the other hand, due to the attention mechanism, the DEformer suffers from the same quadratic complexity problem known to plague Transformers.
However, recent work in sparse Transformers (e.g., \cite{child2019generating, NEURIPS2020_c8512d14, beltagy2020longformer, Kitaev2020Reformer:}) may allow the DEformer to scale to larger inputs.

When training DeepNADE, a mask is randomly generated for each sample by: (1) randomly selecting an integer $c \in \{0, \dots, D - 1\}$ to serve as the number of conditioning variables and (2) randomly assigning a value of one to $c$ locations in the mask and assigning a value of zero to the remaining locations.
The loss for each sample is then:

\[
    \mathcal{L} = \frac{D}{D - c} \sum_{d=1}^{D} (1 - m_{d}) (-\ln{}(f(\hat{\textbf{x}}, \textbf{m})_{(d - 1)C + l_{d}}))
\]

\noindent
where $l_{d}$ is the label for the $d$-th feature of $\textbf{x}$, and $\frac{D}{D - c}$ is a scaling factor ensuring the loss for each sample is an unbiased estimator (which is necessary because the error signal is only computed for $D - c$ features of the sample due to the $1 - m_{d}$ term).
In contrast, for the DEformer, there is always an error signal for all of the features of each sample.
While MADE \cite{germain2015made} also produces an error signal for all of the features of each sample, the authors observed that sampling many different weight masks led to the model underfitting, so it is unclear how well MADE can perform fully OADE.

\subsection{Spatial inputs as feature identities}

A number of neural network architectures operate directly on spatial coordinates, which can be interpreted as feature identities in their various contexts (e.g., images \cite{ha2016generating}, point clouds \cite{pointcloudsurvey}, and 3D scenes \cite{sitzmann2019srns}).
Additionally, \citet{liu2018coord} observed that adding channels to feature maps that contain the spatial coordinates of the pixels greatly improved the performance of convolutional neural networks on certain spatial reasoning tasks.
However, none of these models are performing autoregressive distribution estimation, nor do they employ the interleaved input design of the DEformer.

\section{Conclusion}
In this paper, we described an alternative approach to OADE where the identities of features are included \textit{alongside} their values in the input.
We believe the performance of the DEformer on the binarized-MNIST and POWER datasets is encouraging, and we are excited to see how this architecture can be applied in different contexts.

\section*{Author Contributions}
MAA conceived and implemented the architecture, designed and ran the experiments, and wrote the manuscript.
AN partially funded MAA and provided the GPUs for the experiments.

\section*{Acknowledgements}
We would like to thank Iain Murray and Rasool Fakoor for their helpful feedback.

\bibliography{bib}

\begin{thebibliography}{35}
\providecommand{\natexlab}[1]{#1}
\providecommand{\url}[1]{\texttt{#1}}
\expandafter\ifx\csname urlstyle\endcsname\relax
  \providecommand{\doi}[1]{doi: #1}\else
  \providecommand{\doi}{doi: \begingroup \urlstyle{rm}\Url}\fi

\bibitem[Ahmad et~al.(2017)Ahmad, Lavin, Purdy, and
  Agha]{ahmad2017unsupervised}
Ahmad, S., Lavin, A., Purdy, S., and Agha, Z.
\newblock Unsupervised real-time anomaly detection for streaming data.
\newblock \emph{Neurocomputing}, 262:\penalty0 134--147, 2017.

\bibitem[Alcorn \& Nguyen(2021{\natexlab{a}})Alcorn and
  Nguyen]{alcorn2021baller2vec}
Alcorn, M.~A. and Nguyen, A.
\newblock \texttt{baller2vec}: A multi-entity transformer for multi-agent
  spatiotemporal modeling.
\newblock \emph{arXiv preprint arXiv:2102.03291}, 2021{\natexlab{a}}.

\bibitem[Alcorn \& Nguyen(2021{\natexlab{b}})Alcorn and
  Nguyen]{alcorn2021baller2vecplusplus}
Alcorn, M.~A. and Nguyen, A.
\newblock \texttt{baller2vec++}: A look-ahead multi-entity transformer for
  modeling coordinated agents.
\newblock \emph{arXiv preprint arXiv:2104.11980}, 2021{\natexlab{b}}.

\bibitem[Beltagy et~al.(2020)Beltagy, Peters, and Cohan]{beltagy2020longformer}
Beltagy, I., Peters, M.~E., and Cohan, A.
\newblock Longformer: The long-document transformer.
\newblock \emph{arXiv preprint arXiv:2004.05150}, 2020.

\bibitem[Bulatov(2011)]{bulatov_2011}
Bulatov, Y., Sep 2011.
\newblock URL
  \url{http://yaroslavvb.blogspot.com/2011/09/notmnist-dataset.html}.

\bibitem[Child et~al.(2019)Child, Gray, Radford, and
  Sutskever]{child2019generating}
Child, R., Gray, S., Radford, A., and Sutskever, I.
\newblock Generating long sequences with sparse transformers.
\newblock \emph{arXiv preprint arXiv:1904.10509}, 2019.

\bibitem[Dinh et~al.(2015)Dinh, Krueger, and
  Bengio]{DBLP:journals/corr/DinhKB14}
Dinh, L., Krueger, D., and Bengio, Y.
\newblock {NICE:} non-linear independent components estimation.
\newblock In Bengio, Y. and LeCun, Y. (eds.), \emph{3rd International
  Conference on Learning Representations, {ICLR} 2015, San Diego, CA, USA, May
  7-9, 2015, Workshop Track Proceedings}, 2015.
\newblock URL \url{http://arxiv.org/abs/1410.8516}.

\bibitem[Dinh et~al.(2017)Dinh, Sohl{-}Dickstein, and Bengio]{dinh2017realNVP}
Dinh, L., Sohl{-}Dickstein, J., and Bengio, S.
\newblock Density estimation using real {NVP}.
\newblock In \emph{5th International Conference on Learning Representations,
  {ICLR} 2017, Toulon, France, April 24-26, 2017, Conference Track
  Proceedings}. OpenReview.net, 2017.
\newblock URL \url{https://openreview.net/forum?id=HkpbnH9lx}.

\bibitem[Durkan et~al.(2019)Durkan, Bekasov, Murray, and
  Papamakarios]{durkan2019neural}
Durkan, C., Bekasov, A., Murray, I., and Papamakarios, G.
\newblock Neural spline flows.
\newblock \emph{Advances in Neural Information Processing Systems},
  32:\penalty0 7511--7522, 2019.

\bibitem[Fakoor et~al.(2020)Fakoor, Chaudhari, Mueller, and
  Smola]{fakoor2020trade}
Fakoor, R., Chaudhari, P., Mueller, J., and Smola, A.~J.
\newblock Trade: Transformers for density estimation.
\newblock \emph{Invertible Neural Networks, Normalizing Flows, and Explicit
  Likelihood Models Workshop}, 2020.

\bibitem[Germain et~al.(2015)Germain, Gregor, Murray, and
  Larochelle]{germain2015made}
Germain, M., Gregor, K., Murray, I., and Larochelle, H.
\newblock Made: Masked autoencoder for distribution estimation.
\newblock In Bach, F. and Blei, D. (eds.), \emph{Proceedings of the 32nd
  International Conference on Machine Learning}, volume~37 of \emph{Proceedings
  of Machine Learning Research}, pp.\  881--889, Lille, France, 07--09 Jul
  2015. PMLR.
\newblock URL \url{http://proceedings.mlr.press/v37/germain15.html}.

\bibitem[Guo et~al.(2020)Guo, Wang, Hu, Liu, Liu, and
  Bennamoun]{pointcloudsurvey}
Guo, Y., Wang, H., Hu, Q., Liu, H., Liu, L., and Bennamoun, M.
\newblock Deep learning for 3d point clouds: A survey.
\newblock \emph{IEEE Transactions on Pattern Analysis and Machine
  Intelligence}, pp.\  1--1, 2020.
\newblock \doi{10.1109/TPAMI.2020.3005434}.

\bibitem[Ha(2016)]{ha2016generating}
Ha, D.
\newblock Generating large images from latent vectors.
\newblock \emph{blog.otoro.net}, 2016.
\newblock URL
  \url{https://blog.otoro.net/2016/04/01/generating-large-images-from-latent-vectors/}.

\bibitem[Huang et~al.(2018)Huang, Krueger, Lacoste, and
  Courville]{huang2018neural}
Huang, C.-W., Krueger, D., Lacoste, A., and Courville, A.
\newblock Neural autoregressive flows.
\newblock In \emph{International Conference on Machine Learning}, pp.\
  2078--2087. PMLR, 2018.

\bibitem[Irie et~al.(2019)Irie, Zeyer, Schlüter, and Ney]{Irie2019}
Irie, K., Zeyer, A., Schlüter, R., and Ney, H.
\newblock Language modeling with deep transformers.
\newblock In \emph{Proc. Interspeech 2019}, pp.\  3905--3909, 2019.
\newblock \doi{10.21437/Interspeech.2019-2225}.
\newblock URL \url{http://dx.doi.org/10.21437/Interspeech.2019-2225}.

\bibitem[Jain et~al.(2020)Jain, Abbeel, and Pathak]{jain2020lmconv}
Jain, A., Abbeel, P., and Pathak, D.
\newblock Locally masked convolution for autoregressive models.
\newblock In Peters, J. and Sontag, D. (eds.), \emph{Proceedings of the 36th
  Conference on Uncertainty in Artificial Intelligence (UAI)}, volume 124 of
  \emph{Proceedings of Machine Learning Research}, pp.\  1358--1367. PMLR,
  03--06 Aug 2020.
\newblock URL \url{http://proceedings.mlr.press/v124/jain20b.html}.

\bibitem[Kingma \& Ba(2015)Kingma and Ba]{kingma2014adam}
Kingma, D.~P. and Ba, J.
\newblock Adam: A method for stochastic optimization.
\newblock In \emph{International Conference on Learning Representations}, 2015.

\bibitem[Kingma \& Dhariwal(2018)Kingma and Dhariwal]{kingma2018glow}
Kingma, D.~P. and Dhariwal, P.
\newblock Glow: Generative flow with invertible 1x1 convolutions.
\newblock In Bengio, S., Wallach, H., Larochelle, H., Grauman, K.,
  Cesa-Bianchi, N., and Garnett, R. (eds.), \emph{Advances in Neural
  Information Processing Systems}, volume~31. Curran Associates, Inc., 2018.
\newblock URL
  \url{https://proceedings.neurips.cc/paper/2018/file/d139db6a236200b21cc7f752979132d0-Paper.pdf}.

\bibitem[Kingma et~al.(2016)Kingma, Salimans, Jozefowicz, Chen, Sutskever, and
  Welling]{kingma2016iaf}
Kingma, D.~P., Salimans, T., Jozefowicz, R., Chen, X., Sutskever, I., and
  Welling, M.
\newblock Improved variational inference with inverse autoregressive flow.
\newblock In Lee, D., Sugiyama, M., Luxburg, U., Guyon, I., and Garnett, R.
  (eds.), \emph{Advances in Neural Information Processing Systems}, volume~29.
  Curran Associates, Inc., 2016.
\newblock URL
  \url{https://proceedings.neurips.cc/paper/2016/file/ddeebdeefdb7e7e7a697e1c3e3d8ef54-Paper.pdf}.

\bibitem[Kitaev et~al.(2020)Kitaev, Kaiser, and Levskaya]{Kitaev2020Reformer:}
Kitaev, N., Kaiser, L., and Levskaya, A.
\newblock Reformer: The efficient transformer.
\newblock In \emph{International Conference on Learning Representations}, 2020.
\newblock URL \url{https://openreview.net/forum?id=rkgNKkHtvB}.

\bibitem[Krizhevsky et~al.(2009)Krizhevsky, Hinton,
  et~al.]{krizhevsky2009learning}
Krizhevsky, A., Hinton, G., et~al.
\newblock Learning multiple layers of features from tiny images.
\newblock 2009.

\bibitem[Larochelle \& Murray(2011)Larochelle and Murray]{larochelle2011neural}
Larochelle, H. and Murray, I.
\newblock The neural autoregressive distribution estimator.
\newblock In \emph{Proceedings of the Fourteenth International Conference on
  Artificial Intelligence and Statistics}, pp.\  29--37. JMLR Workshop and
  Conference Proceedings, 2011.

\bibitem[Liu et~al.(2018)Liu, Lehman, Molino, Petroski~Such, Frank, Sergeev,
  and Yosinski]{liu2018coord}
Liu, R., Lehman, J., Molino, P., Petroski~Such, F., Frank, E., Sergeev, A., and
  Yosinski, J.
\newblock An intriguing failing of convolutional neural networks and the
  coordconv solution.
\newblock In Bengio, S., Wallach, H., Larochelle, H., Grauman, K.,
  Cesa-Bianchi, N., and Garnett, R. (eds.), \emph{Advances in Neural
  Information Processing Systems}, volume~31. Curran Associates, Inc., 2018.
\newblock URL
  \url{https://proceedings.neurips.cc/paper/2018/file/60106888f8977b71e1f15db7bc9a88d1-Paper.pdf}.

\bibitem[Oord et~al.(2016)Oord, Kalchbrenner, and
  Kavukcuoglu]{oord2016pixelrnn}
Oord, A.~V., Kalchbrenner, N., and Kavukcuoglu, K.
\newblock Pixel recurrent neural networks.
\newblock In Balcan, M.~F. and Weinberger, K.~Q. (eds.), \emph{Proceedings of
  The 33rd International Conference on Machine Learning}, volume~48 of
  \emph{Proceedings of Machine Learning Research}, pp.\  1747--1756, New York,
  New York, USA, 20--22 Jun 2016. PMLR.
\newblock URL \url{http://proceedings.mlr.press/v48/oord16.html}.

\bibitem[Papamakarios et~al.(2017)Papamakarios, Pavlakou, and
  Murray]{papamakarios2017maf}
Papamakarios, G., Pavlakou, T., and Murray, I.
\newblock Masked autoregressive flow for density estimation.
\newblock In Guyon, I., Luxburg, U.~V., Bengio, S., Wallach, H., Fergus, R.,
  Vishwanathan, S., and Garnett, R. (eds.), \emph{Advances in Neural
  Information Processing Systems}, volume~30. Curran Associates, Inc., 2017.
\newblock URL
  \url{https://proceedings.neurips.cc/paper/2017/file/6c1da886822c67822bcf3679d04369fa-Paper.pdf}.

\bibitem[Papamakarios et~al.(2021)Papamakarios, Nalisnick, Rezende, Mohamed,
  and Lakshminarayanan]{Papamakarios2021normalizing}
Papamakarios, G., Nalisnick, E., Rezende, D.~J., Mohamed, S., and
  Lakshminarayanan, B.
\newblock Normalizing flows for probabilistic modeling and inference.
\newblock \emph{Journal of Machine Learning Research}, 22\penalty0
  (57):\penalty0 1--64, 2021.
\newblock URL \url{http://jmlr.org/papers/v22/19-1028.html}.

\bibitem[Salakhutdinov \& Murray(2008)Salakhutdinov and
  Murray]{salakhutdinov2008quantitative}
Salakhutdinov, R. and Murray, I.
\newblock On the quantitative analysis of deep belief networks.
\newblock In \emph{Proceedings of the 25th international conference on Machine
  learning}, pp.\  872--879, 2008.

\bibitem[Salimans et~al.(2017)Salimans, Karpathy, Chen, and
  Kingma]{Salimans2017PixeCNN}
Salimans, T., Karpathy, A., Chen, X., and Kingma, D.~P.
\newblock Pixelcnn++: A pixelcnn implementation with discretized logistic
  mixture likelihood and other modifications.
\newblock In \emph{ICLR}, 2017.

\bibitem[Sitzmann et~al.(2019)Sitzmann, Zollh{\"o}fer, and
  Wetzstein]{sitzmann2019srns}
Sitzmann, V., Zollh{\"o}fer, M., and Wetzstein, G.
\newblock Scene representation networks: Continuous 3d-structure-aware neural
  scene representations.
\newblock In \emph{Advances in Neural Information Processing Systems}, 2019.

\bibitem[Uria et~al.(2014)Uria, Murray, and Larochelle]{uria2014deep}
Uria, B., Murray, I., and Larochelle, H.
\newblock A deep and tractable density estimator.
\newblock In \emph{International Conference on Machine Learning}, pp.\
  467--475. PMLR, 2014.

\bibitem[Uria et~al.(2016)Uria, C{\^o}t{\'e}, Gregor, Murray, and
  Larochelle]{uria2016neural}
Uria, B., C{\^o}t{\'e}, M.-A., Gregor, K., Murray, I., and Larochelle, H.
\newblock Neural autoregressive distribution estimation.
\newblock \emph{The Journal of Machine Learning Research}, 17\penalty0
  (1):\penalty0 7184--7220, 2016.

\bibitem[Vaswani et~al.(2017)Vaswani, Shazeer, Parmar, Uszkoreit, Jones, Gomez,
  Kaiser, and Polosukhin]{vaswani2017attention}
Vaswani, A., Shazeer, N., Parmar, N., Uszkoreit, J., Jones, L., Gomez, A.~N.,
  Kaiser, {\L}., and Polosukhin, I.
\newblock Attention is all you need.
\newblock In \emph{Advances in Neural Information Processing Systems}, pp.\
  5998--6008, 2017.

\bibitem[Vergara et~al.(2012)Vergara, Vembu, Ayhan, Ryan, Homer, and
  Huerta]{vergara2012chemical}
Vergara, A., Vembu, S., Ayhan, T., Ryan, M.~A., Homer, M.~L., and Huerta, R.
\newblock Chemical gas sensor drift compensation using classifier ensembles.
\newblock \emph{Sensors and Actuators B: Chemical}, 166:\penalty0 320--329,
  2012.

\bibitem[Yang et~al.(2019)Yang, Dai, Yang, Carbonell, Salakhutdinov, and
  Le]{yang2019xlnet}
Yang, Z., Dai, Z., Yang, Y., Carbonell, J., Salakhutdinov, R.~R., and Le, Q.~V.
\newblock Xlnet: Generalized autoregressive pretraining for language
  understanding.
\newblock In Wallach, H., Larochelle, H., Beygelzimer, A., d\textquotesingle
  Alch\'{e}-Buc, F., Fox, E., and Garnett, R. (eds.), \emph{Advances in Neural
  Information Processing Systems}, volume~32. Curran Associates, Inc., 2019.
\newblock URL
  \url{https://proceedings.neurips.cc/paper/2019/file/dc6a7e655d7e5840e66733e9ee67cc69-Paper.pdf}.

\bibitem[Zaheer et~al.(2020)Zaheer, Guruganesh, Dubey, Ainslie, Alberti,
  Ontanon, Pham, Ravula, Wang, Yang, and Ahmed]{NEURIPS2020_c8512d14}
Zaheer, M., Guruganesh, G., Dubey, K.~A., Ainslie, J., Alberti, C., Ontanon,
  S., Pham, P., Ravula, A., Wang, Q., Yang, L., and Ahmed, A.
\newblock Big bird: Transformers for longer sequences.
\newblock In Larochelle, H., Ranzato, M., Hadsell, R., Balcan, M.~F., and Lin,
  H. (eds.), \emph{Advances in Neural Information Processing Systems},
  volume~33, pp.\  17283--17297. Curran Associates, Inc., 2020.
\newblock URL
  \url{https://proceedings.neurips.cc/paper/2020/file/c8512d142a2d849725f31a9a7a361ab9-Paper.pdf}.

\end{thebibliography}
\bibliographystyle{icml2021}



\end{document}